*Learning opacity in Stratal Maximum Entropy Grammar*[*]
Aleksei Nazarov (Harvard University) and Joe Pater (University of Massachusetts Amherst)

1. Introduction

(Kiparsky 1971, 1973) draws attention to cases of historical change that suggest that at least some opaque phonological process interactions are difficult to learn. In his work on Stratal Optimality Theory (Stratal OT), he further claims that independent evidence about the stratal affiliation of a process is helpful in learning an opaque interaction (Kiparsky 2000). In this paper, we develop a computationally implemented learner for a weighted constraint version of Stratal OT, and show that for an initial set of test cases, opaque patterns are indeed generally more difficult than their transparent counterparts, and that information about stratal affiliation does make learning of an opaque pattern easier.

Our results support the viability of Stratal OT as a theory of opaque interactions (Bermúdez-Otero 1999, 2003, Kiparsky 2000; see below). However, it is not clear whether a stratal setup is the only source of opaque interactions – in fact, some cases of opacity may stem from other factors (McCarthy 2007). A broader assessment of alternative theories of opacity will not be attempted in this paper; see Jarosz (2016) for work on learning opacity in one of these alternative frameworks (Jarosz 2014).

In the Stratal OT approach to opacity (see also especially Bermúdez-Otero 1999, 2003), grammars are chained together to produce outputs that would not be possible within a single level. For example, the classic case of the opaque interaction between diphthong raising and flapping in Canadian English illustrated in (1), *mitre* can be produced by raising in the first (word level) grammar, and then flapping in the second (phrase level) one. As we will see shortly, the final candidate cannot be produced by a single OT grammar, with the standard set of constraints assumed for this case (though see Pater 2014).

(1)     /maɪtɚ/ – *Grammar 1* → [mʌɪtɚ] – *Grammar 2* → [mʌɪɾɚ]

This chaining of grammars results in a hidden structure problem (Tesar and Smolensky 2000), insofar as only the output of the final grammar, and not any earlier ones, is available in the learning data. Bermúdez-Otero (2003) develops a revision to the constraint demotion algorithm that is specifically tailored to the problem of hidden structure in Stratal OT. In this paper, we instead apply a general approach to learning hidden structure in probabilistic OT to the specific case of chained grammars.

[*] We would like to thank Ricardo Bermúdez-Otero, Paul Boersma, Jeroen Breteler, Ivy Hauser, Jeff Heinz, Coral Hughto, Gaja Jarosz, Marc van Oostendorp, Olivier Rizzolo, Klaas Seinhorst, and Robert Staubs as well as audiences at the 21st Manchester Phonology Meeting, at the University of Massachusetts Amherst, and at the University of Amsterdam for their very helpful feedback on this paper and for stimulating discussion. We also thank the editors of this volume and two anonymous reviewers for their very helpful and useful comments. We are grateful to the National Science Foundation for supporting this work through grants BCS-0813829 and BCS-1424077. All errors are ours.

The variant of probabilistic OT adopted here is Maximum Entropy Grammar (Goldwater & Johnson 2003, Hayes & Wilson 2008). The 'MaxEnt' formalism has a broad basis in applied mathematics, including connectionism (Smolensky 1986); see Johnson (2013) for an overview, as well as Goldwater & Johnson (2003). MaxEnt defines the probability distribution over a candidate set in a conveniently direct way: candidate probabilities are proportional to the exponential of the weighted sum of violations.

Candidate probabilities are illustrated in the tableau in (2), in which weights are indicated under constraint names (which are simplified versions of the ones used for the case study in section 3.2). The weighted sum appears in the column headed '$H$', for Harmony (Smolensky & Legendre 2006). The exponentiation of H appears under $e^H$, and the resulting probability (obtained by normalization of $e^H$ to sum to 1 across all candidates) under $p$.

(2) *Computation of probabilities over output candidates in MaxEnt*

| /maɪtɚ/ | *aɪt 7 | Ident(C) 5 | Ident(V) 1 | *VTV 0 | $H$ | $e^H$ | $p$ |
|---|---|---|---|---|---|---|---|
| a. maɪtɚ | −1 | | | −1 | −7 | 0.001 | 0.002 |
| b. maɪɾɚ | | −1 | | | −5 | 0.007 | 0.018 |
| c. mʌɪtɚ | | | −1 | −1 | −1 | 0.368 | 0.973 |
| d. mʌɪɾɚ | | −1 | −1 | | −6 | 0.002 | 0.007 |

The *aɪt constraint penalises the unraised diphthong [aɪ] before a voiceless consonant, and Ident(V) penalises raising to [ʌɪ]. With *aɪt having greater weight than Ident(V), raising before a voiceless stop, as in (2c), has greater probability than the faithful (2a). *VTV penalises an alveolar stop between vowels, and Ident(C) penalises the change from stop to flap. With Ident(C) weighted higher than *VTV, the probability on flapping candidates (2b) and (2d) is lowered. Candidate (2d) is the correct output in Canadian English, but it has a proper superset of the violations of (2b), since flapping also satisfies *aɪt. No weighting of the constraints will give [mʌɪɾɚ] the highest probability in the tableau. In this respect, the situation is the same as in single-level OT, which cannot make (2d) optimal. In MaxEnt, (2d) can tie with (2b): when the constraints they violate have zero weight, and they are given equal probability (see Pater 2016 for further discussion and references on harmonic bounding and probabilistic weighted constraint theories). In (26) in section 3.2, we will see that [mʌɪɾɚ] can emerge from a second chained grammar with highest probability.

Our way of dealing with the hidden structure problem posed by Stratal OT is an extension of the approach to Harmonic Serialism in Staubs & Pater (2016). The probability that the grammar assigns to an output form is the sum of the probabilities of the derivations that lead to it. The probability of a derivation is the product of the probabilities of each of its component steps. The probability of each step is determined by the weights of the constraints that apply at that level (word or phrase; see section 2.2 for more details). We use a batch learning approach in which the objective is to minimise the divergence between the predictions of the grammar and the learning data, subject to a regularization term that prefers lower weights. This is a standard approach to learning MaxEnt grammars (see e.g. Goldwater & Johnson 2003, Hayes & Wilson 2008).

Jarosz (2006, ms) shows how this approach of considering every possible hidden structure in generating an overt candidate can be applied to phonology in her work on Expectation Maximization with Optimality Theory. Here, we follow Eisenstat (2009), Pater et al. (2012), Johnson et al. (2015), and Staubs & Pater (2016) in adopting a weighted constraint variant of this general approach.

When there is hidden structure, the learner is not guaranteed to converge on the global optimum, that is, a set of weights that best satisfy the objective function (the solution space is not guaranteed to be convex). Instead it may only find a local optimum, a set of weights that is the best within the space that the learner can explore. We provide examples of such local optima in (13-14) and (27-28) in sections 2.3 and 3.3, respectively.

A simple and standard way of finding a global optimum in such cases is to try different starting positions for the learner (initializations), with the hope that one or more will allow the learner to find the global optimum. In the limit, as long as the space from which we are sampling initializations contains points from which there is a path towards an optimal solution, this will allow the learner to find a global optimum. In our test cases, there were always runs of the learner that granted more than 0.5 probability to the overt forms that had highest probability in the learning data, our criterion of success. Our measure of 'ease of learning' for the various patterns we study is the probability that a learner will be successful, over a set of random initializations.

In the next section we provide more details about our grammatical and learning theories, in the context of our first test case, a relatively simple example of opaque tense-lax vowel alternations in French. In section 3 we turn to the opaque interaction of diphthong raising and flapping in Canadian English, and the effect of supplying evidence to the learner of phrasal non-raising. Section 4 provides an exploration of the mechanisms that lead to incorrect learning of (predominantly) opaque patterns. Finally, overall conclusions and directions for further research can be found in section 5.

## 2. Case study I: Southern French tensing/laxing

## 2.1 Data and analysis

As reported by Moreux (1985, 2006), Rizzolo (2002), and Eychenne (2014), certain dialects of Southern French have a synchronic process which tenses mid vowels in open syllables, while laxing them in closed ones (as well as open syllables followed by a schwa syllable; we will disregard this complication, and refer the reader to Selkirk 1978, Moreux 1985, and Rizzolo 2002 for potential explanations of this). This is conventionally called *loi de position* ('law of position').

(3) *Examples of* loi de position *(Eychenne 2014, Rizzolo 2002)*
a. /sel/ → [sɛl] 'salt'
   /se/ → [se] 'knows'
b. /pøʁ/ → [pœʁ] 'fear'
   /pøʁ-ø/ → [pø.ʁø] 'easily frightened'
c. /poʁ/ → [pɔʁ] 'pore'
   /poʁ-ø/ → [po.ʁø] 'porous'

Tensing/laxing is made opaque (is counterbled) by resyllabification across word boundaries, at least in some cases[1], as shown in (4b) below, where the lax vowel [œ] is kept in the surface open syllable [pœ]:

(4) *Opaque interaction between resyllabification and tensing/laxing* (Rizzolo 2002:51)
a. /kɑ̃pøʁ/ → [kɑ̃.pœʁ] 'camper'
b. /kɑ̃pøʁ ɑ̃ʁaʒe/ → kɑ̃.**pœʁ** # ɑ̃.ʁa.ʒe → [kɑ̃.**pœ**.ʁɑ̃.ʁa.ʒe] 'enraged camper'

In this simulation, we investigated whether our learner could deal with this simple case of opacity. This also serves as a simple case to illustrate in more detail the functioning of the learner. The data that we offered to the learner consisted of exceptionless *loi de position* as well as exceptionless opacity through resyllabification:

(5) *Opaque interaction, as in actual Southern French*
a. /sɛt a/ → (sɛt. # a →) [sɛ.ta] 'this (letter) A' <cette A>
b. /se ta/ → (se. # ta →) [se.ta] 'it is "ta"' <c'est 'ta'>

The constraint set that we used for this subset of real French was maximally simple, as shown in (6). Correct syllabification was taken for granted in order to keep the constraint and candidate sets as small as possible so that the interactions yielding opacity are easy to see.

(6) *Constraint set used for the Southern French case study*
a. *[-tense]/Open : One violation mark for every [-tense] segment in an open syllable.
b. *[+tense]/Closed : One violation mark for every [+tense] segment in a closed syllable.
c. Ident(V) : One violation mark for every vowel that is not identical to its input correspondent.

The original formulation of Stratal OT (Kiparsky 2000, Bermudéz-Otero 1999) allows for three levels of derivation: a stem level, a word level, and a phrase level – which all have the same constraints, but different rankings or weightings. For this problem, we will only consider two levels – a word level and a phrase level grammar. The word level grammar evaluates each word individually, without regard to its neighbors in the phrase. By contrast, the phrase level, which operates on the output of the word level grammar, does evaluate entire phrases together.

  With this setup, then, the opaque interaction is obtained when word level *[+tense]/Closed and *[-tense]/Open have high weight and word level Ident(V) has low weight, while the opposite holds at the phrase level. This corresponds to activity of the *loi de position* at the word level, and its inactivity at the phrase level.

  This latter scenario is the only option to derive this opaque interaction. As can be seen in tableau (8) below, phrase level Markedness disprefers surface [sɛ.ta], so that the mapping from /e/ to [ɛ] cannot be derived at the phrase level. At the same time, word level Markedness prefers the word level output that does map /e/ to [ɛ], sɛt # a, because the first vowel is in a closed syllable at that level, since the phrasal context is invisible. This can be seen in tableau (7a) below.

---

[1] Rizzolo (2002:51) reports that this effect is not unexceptional, at least in the dialect he describes. In our case study, we will investigate an idealised version of this pattern in which the opaque interaction is unexceptional.

Activity of *loi de position* at the word level can be expressed with the following weights (found at successful runs of the learner described in section 2.2):

(7) *Word level tableaux for the data in* (5)

| a. /set#a/ | *[-tense]/Open 6.24 | *[+tense]/Cl 6.24 | Ident(V) 0 | H | $e^H$ | *p* |
|---|---|---|---|---|---|---|
| set # a | | −1 | | −6.24 | 0.002 | 0.00 |
| sɛt # a | | | −1 | 0 | 1 | 1.00 |

| b. /se#ta/ | *[-tense]/Open 6.24 | *[+tense]/Cl 6.24 | Ident(V) 0 | H | $e^H$ | *p* |
|---|---|---|---|---|---|---|
| se # ta | | | | 0 | 1 | 1.00 |
| sɛ # ta | −1 | | −1 | −6.24 | 0.002 | 0.00 |

At the phrase level, giving high weight to Ident(V) and zero weight to both Markedness constraints results in 1.00 probability for all faithful mappings. This is illustrated in (9) for the phrase level derivation that takes place if the word level form output is sɛt. # a: phrase level outputs retain the [ɛ] that was created by closed syllable tensing at the word level with a probability of 1.00.

(8) *Phrase level tableau for* (5a)

| /sɛt # a/ | *[-tense]/Open 0 | *[+tense]/Cl 0 | Ident(V) 6.93 | H | $e^H$ | p |
|---|---|---|---|---|---|---|
| se.ta | | | −1 | −6.93 | 0.001 | 0.00 |
| sɛ.ta | −1 | | | 0 | 1 | 1.00 |

These single-stratum mappings shown in (7-8) are assembled into derivational paths that lead from an underlying representation (UR) to a surface representation (SR). The probability of a single derivational path is obtained by multiplying the probabilities of every step in the path, as illustrated in (9). This is because every derivational step in Stratal OT is, by definition, independent from earlier or later derivational steps (Bermúdez-Otero 1999, Kiparsky 2000)[2].

(9) *Probabilities of derivational paths*
a. p(/set#a/ → set # a → [se.ta]) =
   p(/set#a/ → set # a) x p(set # a → [se.ta]) = 0.00 x 1.00 = 0.00

b. p(/set#a/ → sɛt # a → [sɛ.ta]) =
   p(/set#a/ → sɛt # a) x p(sɛt # a → [sɛ.ta]) = 1.00 x 1.00 = 1.00

---

[2] See also Odden (2011) on the Markov property traditionally ascribed to ordered phonological rules.

The expected probability of a surface form (more accurately, a UR/SR mapping) is the sum of the probabilities of all derivational paths that lead from the input to that surface form, as in (10).[3] This table shows that, given the weights in (7-8), the desired output candidates (cf. (5)) are generated with a probability of 1.00.

(10) *Expected probabilities of UR/SR mappings: sum over all derivational paths*

| Underlying | Derivational path | Probability | Surface | Probability |
|---|---|---|---|---|
| /set#a/ | /set#a/ → set # a → [se.ta] | 0.00 | [se.ta] | 0.00 |
| | /set#a/ → sɛt # a → [se.ta] | 0.00 | | |
| | /set#a/ → set # a → [sɛ.ta] | 0.00 | [sɛ.ta] | 1.00 |
| | /set#a/ → sɛt # a → [sɛ.ta] | 1.00 | | |
| /se#ta/ | /se#ta/ → se # ta → [se.ta] | 1.00 | [se.ta] | 1.00 |
| | /se#ta/ → sɛ # ta → [se.ta] | 0.00 | | |
| | /se#ta/ → se # ta → [sɛ.ta] | 0.00 | [sɛ.ta] | 0.00 |
| | /se#ta/ → sɛ # ta → [sɛ.ta] | 0.00 | | |

Now that we have explained and illustrated the generation of probabilities over UR/SR mappings given the successful weights in (7,8), we will explain the structure of the learner that arrived at these weights in section 2.2. Then, in section 2.3, we will show how often this learner arrives at this successful set of weights within a sample of 100 runs, and what happens within that same sample when the learner does not find this analysis.

## 2.2 Learning

Staubs (2014b) provides an implementation in *R* (R Core Team 2013) of a variant of Staubs & Pater's (2016) approach to learning serial grammars. We modified it minimally to allow different violations of the same constraint at different derivational stages, which is sometimes necessary in a Stratal framework.

For instance, the constraint *[+tense]/Closed in our Southern French simulation will be violated at the word level by a sequence [set # a], because even through [t] and [a] could form a separate syllable, the word boundary between them prevents this. However, the same constraint remains unviolated at the phrase level for the same sequence, since the grammar can now look beyond word boundaries and evaluate entire phrases, so that [t] and [a] do form their own syllable: [se.ta].

Expected probabilities of UR/SR mappings are computed given sets of candidates, violation vectors, and constraint weights, as described and illustrated in section 2.1.

---

[3] Since there are several derivational paths per surface form, but only one surface form per derivational path, UR/SR mappings are bundles of non-overlapping sets of derivational paths. This means that, in computing expected probabilities of UR/SR mappings, each derivational path's probability is counted once, so that UR/SR probabilities are guaranteed to add up to 1.

In addition, the learner is also given observed probabilities of UR/SR mappings are provided to the learner. Since we are working with categorical data, all UR/SR mappings found in the data were given a probability of 1 on all UR/SR mappings found in the data, and all others were given a probability of 0 on all others. The learner then minimises the Kullback-Leibler-divergence (KL-divergence; Kullback & Leibler 1951) of the expected probabilities from the observed probabilities, which is a measure of how closely the grammar has been able to fit the data.

(11) *KL-divergence*
For all UR/SR mappings UR→SR$_{i=1}$, ..., UR→SR$_{i=k}$:

$$D_{KL}(P_{obs} \parallel P_{exp}) = \sum_{i=1}^{k} p_{obs}(UR \to SR_i) \times \log\left(\frac{p_{obs}(UR \to SR_i)}{p_{exp}(UR \to SR_i)}\right)$$

Minimization of this objective function is done by the L-BFGS-B method of optimization (Byrd et al. 1995). This is a pseudo-Newtonian batch optimization method which uses approximations of the first-order and second-order derivatives of the objective function to take incremental steps towards an optimum. The 'optim' implementation of this method in *R* was used, specifying a minimum of zero on all constraint weights.

To place an upper bound on the objective function, an L2 regularisation term (Hoerl & Kennard 1970) with $\mu = 0$ and $\sigma^2 = 10{,}000$ is added to it. This regularisation term penalises weights as they increase, keeping the optimal solution finite, and driving a constraint's weight down to zero whenever it is not helpful in the analysis.

Summarising, our algorithm calculates probabilities of derivations in Stratal MaxEnt, sums over them to get the expected probabilities of overt forms, and fits the observed distribution of UR/SR mappings by minimising KL-divergence.

2.3 Results

The learner described in section 2.2 was run 100 times on the French *loi de position* data, with every run starting from random starting weights for each constraint, drawn i.i.d. from a uniform distribution over [0, 10]. The results are as follows:

(12) *Results for the Southern French data set*

| Data set | Learned successfully out of 100 runs |
|---|---|
| /set#a/ → [sɛ.ta] <br> /se#ta/ → [se.ta] | 96 |

Thus, opaque *loi de position* was learned successfully for an overwhelming majority of start weights. However, there is still a slight chance of incorrect learning, which we defined as a probability of 0.5 or less on at least one surface form that has a probability of 1 in the learning data (see also section 1). The 4 unsuccessful runs yielded grammars which had free variation between tense [e] and lax [ɛ]:

(13) *UR/SR mapping probabilities for the local optimum*

| /set#a/ | probability | /se#ta/ | probability |
|---|---|---|---|
| [se.ta] | 0.50 | [se.ta] | 0.50 |
| [sɛ.ta] | 0.50 | [sɛ.ta] | 0.50 |

This is a local optimum that arises when all constraint weights are set to zero, as in (14). KL-divergence from the actual data (our measure of error) is smaller for these weights than for any other weights within the weight space explored by the learner.

(14) *Weights that generate the local optimum*

| Word level | | | Phrase level | | |
|---|---|---|---|---|---|
| *[-tense]/Open | *[+tense]/Cl | Ident(V) | *[-tense]/Open | *[+tense]/Cl | Ident(V) |
| 0 | 0 | 0 | 0 | 0 | 0 |

As will be explained in section 4.1, an initially incorrect weighting at the word level leads to a tendency for phrase level Faithfulness to lower its weight. At the same time, when phrase level Ident(V) reaches zero weight, any effect of closed syllable laxing from the word level cannot be transmitted to the final output. Once this happens, 0.50 probability for all candidates is the best possible fit to the data, and the all-zero weight version of that solution minimises the penalty on the regularization term, which prefers smaller weights.

It could be argued that this weight set is not a local optimum in the strict sense. There is a locally available solution that is better than all-zero weights: assigning minimal weight (0+ε for any positive ε) to word level *[-tense]/Open and *[+tense]/Closed as well as phrase level Ident(V), while keeping all other constraints at 0, will lower the objective function with respect to all-zero weights, as this will nudge the probabilities of both /set#a/ → [sɛ.ta] and /se#ta/ → [se.ta] slightly above 0.50. However, for the reasons outlined above, the learner cannot see this possibility when it retreats to all-zero weights.

## 3 Case study II: Raising and flapping in Canadian English

### 3.1 Data

Our second case study is a classic case of opacity, attested in Canadian English (Joos 1942, Idsardi 2000 and references therein, Pater 2014): the interaction between Canadian Raising and flapping. Low-nucleus diphthongs [aɪ, aʊ] raise to [ʌɪ, ʌʊ] before voiceless consonants, as in (15a), and coronal oral stops /t, d/ become a voiced flap [ɾ] in a variety of contexts (De Jong 2011) – among others, between two vowels if the first is stressed and the second is not, as in (15b).

The two processes interact in a counterbleeding way: the fact that flapping cancels the voicelessness of underlying /t/ does not prevent that /t/ from triggering raising. This is illustrated in (15c).

(15) *Canadian English counterbleeding interaction*
a. /laɪf/ → [lʌɪf] 'life'    cf. /laɪ/ → [laɪ] 'lie'
b. /kʌt-ɚ/ → [ˈkʌɾɚ] 'cutter'
c. /maɪtɚ/ → [mʌɪɾɚ] 'mitre' cf. /saɪdɚ/ → [saɪɾɚ] 'cider'

In addition, raising is restricted to the word domain, as illustrated in (16a), while there is no evidence of such a restriction for flapping, as in (16b). This fits with a stratal analysis in which raising applies at the word level only, while flapping applies at the phrase level only – as illustrated in (17).

(16) *Word-boundedness of raising*
a. /laɪ#fɔɹ/ → [laɪ fɔɹ] 'lie for'     *[lʌɪ fɔɹ], cf. [lʌɪf]
b. /laɪ#tu/ → [laɪ ɾə] 'lie to'        *[lʌɪ ɾə], cf. [mʌɪɾɚ]

(17) *Sketch of stratal analysis of flapping and raising*
/maɪtɚ/         →      mʌɪtɚ           →      mʌɪɾɚ
Underlying             Word Level:             Phrase Level:
Representation         Raising only            Flapping only

The transparent counterpart to this pattern, claimed to be spoken by some Canadian English speakers (Joos 1942), is a language which also has both raising and flapping as in (15ab) and (16), but the application of flapping blocks the application of raising, as in (18), since the flap [ɾ] is not voiceless. However, the existence of this transparent dialect has been disputed in later literature (Kaye 1990). In our simulations, we use this (perhaps hypothetical) transparent dialect as a non-opaque baseline against which to compare the mainstream, opaque dialect of Canadian English.

(18) *Transparent interaction between raising and flapping*
/maɪtɚ/ → **[maɪɾɚ]** 'mitre'

In the next subsection, we will investigate the learnability of the opaque and the transparent versions of the Canadian English data. We will consider the contrast between opacity and transparency, and between various possible datasets.

## 3.2 Simulation setup

Canadian English provides at least two pieces of independent evidence regarding the stratal affiliation of the opaque raising process: evidence for its application outside the flapping context, as in (15a), and evidence for its word-boundedness, as in (16a). Based on Kiparsky (2000), we predict the availability of such evidence to make the opaque interaction more learnable. To investigate this, we considered four data sets representing various degrees of independent evidence for the stratal affiliation of the opaque process.

'Mitre-cider' has no independent evidence regarding the opaque process – its synchronic activity and stratal affiliation must be inferred from the opaque interaction. 'Mitre-cider-life' has evidence for the opaque process' applying outside of the interaction, since the [f] in 'life' is not subject to flapping: raising must be synchronically active at some stratum. 'Mitre-cider-lie-for' provides evidence for the non-application of raising at the phrase level, since 'lie for' does not undergo raising despite having [f] after the diphthong across a word boundary. Finally, 'mitre-cider-life-lie-for' combines both pieces of evidence. These data sets were investigated with both an opaque and a transparent interaction between raising and flapping.

(19) *Opaque datasets for Canadian English*

|  | 'mitre-cider' | 'mitre-cider-life' | 'mitre-cider-lie-for' | 'mitre-cider-life-lie-for' |
|---|---|---|---|---|
| Opaque | /maɪtɚ/ → [mʌɪɾɚ]<br>/saɪdɚ/ → [saɪɾɚ] | /maɪtɚ/ → [mʌɪɾɚ]<br>/saɪdɚ/ → [saɪɾɚ]<br>/laɪf/ → [lʌɪf] | /maɪtɚ/ → [mʌɪɾɚ]<br>/saɪdɚ/ → [saɪɾɚ]<br><br>/laɪ#fɔɹ/ → [laɪ fɔɹ] | /maɪtɚ/ → [mʌɪɾɚ]<br>/saɪdɚ/ → [saɪɾɚ]<br>/laɪf/ → [lʌɪf]<br>/laɪ#fɔɹ/ → [laɪ fɔɹ] |
| Transparent | /maɪtɚ/ → [maɪɾɚ]<br>/saɪdɚ/ → [saɪɾɚ] | /maɪtɚ/ → [maɪɾɚ]<br>/saɪdɚ/ → [saɪɾɚ]<br>/laɪf/ → [lʌɪf] | /maɪtɚ/ → [maɪɾɚ]<br>/saɪdɚ/ → [saɪɾɚ]<br><br>/laɪ#fɔɹ/ → [laɪ fɔɹ] | /maɪtɚ/ → [maɪɾɚ]<br>/saɪdɚ/ → [saɪɾɚ]<br>/laɪf/ → [lʌɪf]<br>/laɪ#fɔɹ/ → [laɪ fɔɹ] |

Our learner had access to the four constraints in (20). As in the Southern French case, only two derivational levels were used: a word level and a phrase level.

Whenever *aɪ,aʊ/_[-voice] is sufficiently higher weighted than Ident(low), the raising process will be in effect. The flapping process is active when *V́TV is sufficiently above Ident(sonorant)[4].

(20) *Constraint set for Canadian Raising simulations*
a. Ident(low) : One violation mark for raising or lowering a diphthong.
b. Ident(sonorant) : One violation mark for any underlying consonant whose [±sonorant] specification is not identical to that of its output correspondent. This constraint penalises the transition from /t, d/ to flap, or *vice versa*.
c. *V́TV : One violation mark for an alveolar stop [t, d] in between two vowels of which the first is stressed and the second is not.
d. *aɪ,aʊ/_[-voice] : One violation mark for a non-raised diphthong before a voiceless consonant.

The opaque interaction is captured if there is raising but no flapping at the word level, and flapping but no raising at the phrase level (see section 1 on the impossibility of capturing the interaction within one stratum). This is reflected in the weights found at successful learning trials for 'mitre-cider-life-lie-for'. At the word level, *aɪ,aʊ/_[-voice] is far above Ident(low), and *V́TV is far below Ident(sonorant); the opposite is true of the phrase level:

---

[4] We assume that the transition from /t,d/ to [ɾ] entails a change in [±sonorant].

(21) *Sample successful weights for opaque 'mitre-cider-life-lie-for'*

| Word level | | | | Phrase level | | | |
|---|---|---|---|---|---|---|---|
| Ident (son) | Ident (low) | *V́TV | *aɪ,aʊ/_ [-vce] | Ident (son) | Ident (low) | *V́TV | *aɪ,aʊ/_ [-vce] |
| 10.44 | 5.02 | 0 | 11.13 | 0 | 6.81 | 6.12 | 0 |

The word level tableau in (22) below shows that the word level weights above ensure raising but no flapping at the word level.

(22) *Word level tableau for 'mitre'*

| /maɪtɚ/ | Ident(son) 10.44 | Ident(low) 5.02 | *V́TV 0 | *aɪ,aʊ/_[-vce] 11.13 | H | e^H | p |
|---|---|---|---|---|---|---|---|
| maɪtɚ |  |  | −1 | −1 | −11.13 | 0.00001 | 0.00 |
| maɪɾɚ | −1 |  |  |  | −10.44 | 0.00003 | 0.00 |
| mʌɪtɚ |  | −1 | −1 |  | −5.02 | 0.007 | 0.99 |
| mʌɪɾɚ | −1 | −1 |  |  | −15.46 | 0.00000 | 0.00 |

The candidates considered at the phrase level are the same as the candidate set at the word level.[6] As can be seen in the phrase level tableaux in (23), it is essential for the generation of these probabilities that /maɪtɚ/ exhibit raising at the word level, since the phrase level must preserves diphthongs faithfully, or else /maɪtɚ/ will not come out with both raising and flapping (i.e., as attested [mʌɪɾɚ]).

(23) *Two of the phrase level tableaux for 'mitre'*

| a. mʌɪtɚ | Ident(son) 0 | Ident(low) 6.81 | *V́TV 6.12 | *aɪ,aʊ/_[-vce] 0 | H | e^H | p |
|---|---|---|---|---|---|---|---|
| [maɪtɚ] |  | −1 | −1 | −1 | −12.93 | 0.000 | 0.00 |
| [maɪɾɚ] | −1 | −1 |  |  | −6.81 | 0.001 | 0.00 |
| [mʌɪtɚ] |  |  | −1 |  | −6.12 | 0.002 | 0.00 |
| [mʌɪɾɚ] | −1 |  |  |  | 0 | 1 | 1.00 |

| b. maɪɾɚ | Ident(son) 0 | Ident(low) 6.81 | *V́TV 6.12 | *aɪ,aʊ/_[-vce] 0 | H | e^H | p |
|---|---|---|---|---|---|---|---|
| [maɪtɚ] | −1 |  | −1 | −1 | −6.12 | 0.002 | 0.00 |
| [maɪɾɚ] |  |  |  |  | 0 | 1 | 1.00 |
| [mʌɪtɚ] | −1 | −1 | −1 |  | −12.93 | 0.000 | 0.00 |
| [mʌɪɾɚ] |  | −1 |  |  | −6.81 | 0.001 | 0.00 |

---

[6] We did not consider derivational paths with the changes /t/→ɾ→ [d] or /d/→ɾ→ [t]: the phrase level candidate set did not include [m{a,ʌ}idɚ] or [s{a,ʌ}itɚ], so that every tableau had four candidates. However, we are confident that the presence of these candidates would not have significantly changed our results: as shown in (23b), candidates that change ɾ to [t] or [d] at the phrase level incur extra violations of Ident(son) and *V́TV, while not yielding any improvement on the other two constraints.

The surface form probabilities resulting from the weights in (21) match the actual opaque Canadian English data extremely closely, as shown in (24).

(24) *Surface form probabilities generated by* (21)

| /maɪtɚ/ | probability | /saɪdɚ/ | probability | /laɪf/ | probability | /laɪ#fɔɹ/ | probability |
|---|---|---|---|---|---|---|---|
| [maɪtɚ] | 0.00 | [saɪdɚ] | 0.00 | [laɪf] | 0.00 | [laɪ fɔɹ] | **0.99** |
| [maɪɾɚ] | 0.00 | [saɪɾɚ] | **0.99** | | | | |
| [mʌɪtɚ] | 0.00 | [sʌɪdɚ] | 0.00 | [lʌɪf] | **1.00** | [lʌɪ fɔɹ] | 0.01 |
| [mʌɪɾɚ] | **0.99** | [sʌɪɾɚ] | 0.00 | | | | |

As opposed to the opaque interaction, the transparent interaction does not need to time flapping after raising. In fact, for transparent 'mitre-cider' and 'mitre-cider-lie-for', raising need not be represented in the grammar at all, because the raised diphthong is absent from the data – see the lack of weight on *aɪ,aʊ/_[-voice] at either level in (25a-b).

For transparent 'mitre-cider-life', raising can be represented either at the word level, by giving high weight to word level *aɪ,aʊ/_[-voice] and zero weight to word level Ident(sonorant), as in (25c) – or at the phrase level by giving high weight to *V́TV and *aɪ,aʊ/_[-voice] at the phrase level, as in (25d).

Transparent 'mitre-cider-life-lie-for', however, does require raising to take place at the word level, because the non-raised diphthong in /laɪ#fɔɹ/ → [laɪ fɔɹ] precludes raising at the phrase level. The weights found for this dataset are very similar to the second set of weights for 'mitre-cider-life', as shown in (25e).

(25) *Sample weights for successful runs of various transparent datasets*

|   | Dataset (transparent) | Word level | | | | Phrase level | | | |
|---|---|---|---|---|---|---|---|---|---|
|   |   | Ident (son) | Ident (low) | *V́TV | *aɪ,aʊ/_ [-vce] | Ident (son) | Ident (low) | *V́TV | *aɪ,aʊ/_ [-vce] |
| a. | 'mitre-cider' | 0 | 6.78 | 0 | 0 | 0 | 6.78 | 6.78 | 0 |
| b. | 'mitre-cider-lie-for' | 0 | 7.15 | 0 | 0 | 0 | 7.15 | 6.75 | 0 |
| c. | 'mitre-cider-life', var. 1 | 0 | 6.41 | 0 | 0 | 0 | 5.72 | 5.73 | 11.45 |
| d. | 'mitre-cider-life', var. 2 | 0 | 6.07 | 0 | 11.74 | 0 | 6.77 | 6.36 | 0 |
| e. | 'mitre-cider-life-lie-for' | 0 | 6.34 | 0 | 11.98 | 0 | 7.03 | 6.34 | 0 |

Thus, the opaque interaction of raising and flapping requires raising at the word level and flapping at the phrase level. The transparent interaction, however, only requires flapping at the word level, while raising can be represented at either level.

Finally, the addition of both 'life' and 'lie for' to the transparent data set leads to the necessity of representing raising at the word level, even though this is not required for the transparent interaction otherwise.

## 3.3 Results

Simulations were run as described in section 2.2, except that L2 regularization was made stronger by setting $\sigma^2$ to 9,000 instead of 10,000 to prevent the learner from considering constraint weights that tend towards infinity for the opaque 'mitre-cider' dataset.

All four data sets were examined with both opaque and transparent interaction between raising and flapping. The same 100 sets of initializations drawn i.i.d. from a uniform distribution over [0,10] were used for all 8 datasets. The results are as follows:

(26) *Results for Canadian English, for 100 sets of initializations*

| Dataset | Opaque: learned successfully out of 100 runs | Transparent: learned successfully out of 100 runs |
|---|---|---|
| 'mitre-cider' | 51 | 100 |
| 'mitre-cider-life' | 61 | 99 |
| 'mitre-cider-lie-for' | 87 | 100 |
| 'mitre-cider-life-lie-for' | 92 | 93 |

As can be seen in (29), independent evidence regarding the opaque process yields a clear increase in learnability for the opaque cases. Evidence that the opaque process is word-bounded ('lie for') has a stronger effect than evidence for independent activity of the opaque process ('life').

For all transparent datasets except 'mitre-cider-life-lie-for', performance is (almost) at ceiling. For 'mitre-cider-life-lie-for', however, the opaque and transparent versions are learned at a near equal rate.

Whenever (opaque or transparent) Canadian English was not learned successfully (i.e., at least one attested SR was given a probability of 0.5 or less by the grammar), the learner ended up in one of the 4 local optima that are summarised in (27). The table lists, for each underlying representation, the surface representations that have more than 0.00 probability in that local optimum. The symbol '~' will be used as a shorthand for 0.50 probability on both surface representations shown, unless indicated otherwise. Weights that generate each local optimum are given in (28).

(27) *Local optima found for Canadian English simulations*

| Optimum | Occurs in: | Inputs | Outputs |
|---|---|---|---|
| I | **Opaque** 'mitre-cider', 'mitre-cider-lie-for', 'mitre-cider-life-lie-for' | /maɪtɚ/ /saɪdɚ/ (/laɪf/) (/laɪ#fɔɹ/) | [mʌɪɾɚ]~[maɪɾɚ] [sʌɪɾɚ]~[saɪɾɚ] ([laɪf]~[lʌɪf]) ([laɪ fɔɹ]~[lʌɪ fɔɹ]) |
| II | **Opaque** 'mitre-cider-lie-for', 'mitre-cider-life-lie-for' | /maɪtɚ/ /saɪdɚ/ (/laɪf/) /laɪ#fɔɹ/ | [mʌɪɾɚ] 0.67 ~ [maɪɾɚ] [sʌɪɾɚ] 0.67 ~ [saɪɾɚ] ([lʌɪf]) [laɪ fɔɹ] 0.67 ~ [lʌɪ fɔɹ] |
| III | **Opaque** 'mitre-cider-life' **Transparent** 'mitre-cider-life' | /maɪtɚ/ /saɪdɚ/ /laɪf/ - | [mʌɪɾɚ]~[maɪɾɚ] [sʌɪɾɚ]~[saɪɾɚ] [lʌɪf] |
| IV | **Transparent** 'mitre-cider-life-lie-for' | /maɪtɚ/ /saɪdɚ/ /laɪf/ /laɪ#fɔɹ/ | [maɪɾɚ] [saɪɾɚ] [laɪf]~[lʌɪf] [laɪ fɔɹ]~[lʌɪ fɔɹ] |

(28) *Sample weights for local optima*

|  | Word level | | | | Phrase level | | | |
|---|---|---|---|---|---|---|---|---|
|  | Ident (son) | Ident (low) | *V́TV | *aɪ,aʊ/_ [-vce] | Ident (son) | Ident (low) | *V́TV | *aɪ,aʊ/_ [-vce] |
| Local optimum I | 0 | 0 | 0 | 0 | 0 | 0 | 7.75 | 0 |
| Local optimum II | 0 | 6.60 | 0 | 0 | 0 | 5.91 | 5.92 | 5.90 |
| Local optimum III | 0 | 0.69 | 6.14 | 6.54 | 6.51 | 5.86 | 0.04 | 0 |
| Local optimum IV | 0 | 0 | 0 | 0 | 0 | 0 | 7.01 | 6.60 |

These four local optima have in common that they try to represent both processes with minimal appeal to the interaction between levels. For instance, local optima I and IV are attempts to represent the data without appealing to underlying representations, by setting all Faithfulness constraints to zero.

Local optimum II does not represent the raising process at the word level, as necessary for any data set that involves 'life' with a raised diphthong and 'lie for' with a non-raised diphthong (see section 3.2). This means that the vowel contrast between 'life' and 'lie for' has to be modeled as within-word variation.

Finally, local optimum III is a consequence of representing both raising and flapping at the word level. Since applying raising and flapping at the same level leads to lack of raising in 'mitre', variation between raising and non-raising is created by lowering the weight of word level Ident(low).

As in the case of Southern French, local optima I and IV are not a local optima in the narrow sense: if the weights of word level Ident(son), Ident(low) (in the case of local optimum I), and **aɪ,aʊ/_[-vce] and phrase level Ident(low) are increased by even a little bit, this will decrease the value of the objective function. It is the 'bottleneck effect' (see section 4.1) that leads the learner to land in this state. However, local optima II and III are true local optima in the sense that any neighboring values will increase the value of the objective function.

As a final note, none of these outcomes represent opaque analyses. Local optima II and III resemble opaque analyses, since they both assign a significant amount of probability to [mʌɪɾɚ], but they crucially ignore the difference between underlying /t/ in 'mitre' and underlying /d/ in 'cider' and allow unmotivated raising in both words (because the constraints that regulate raising are either at 0 or tied with other constraints).

## 3.4 Summary

Summarising, we have found that independent evidence about the opaque pattern's stratal affiliation can significantly improve the learnability of the opaque interaction – especially the addition of evidence that the opaque process does not apply at the phrase level. Furthermore, the presence of both 'lie' and 'lie for' made the transparent and the opaque interaction equally learnable, in fact destroying the learning advantage of the transparent interaction over the opaque one.

Whenever the languages are not learned successfully, either phrase level Faithfulness is given zero weight, making it impossible to transfer information from the word level to the phrase level, or raising and flapping are represented at the same level when they need to be represented at different levels. We will now turn to a discussion of what we call the 'bottleneck effect', an obstacle posed by the current hidden structure learning problem that leads to the learner's landing in local optima, as well as ways in which evidence for stratal affiliation helps the learner overcome this obstacle.

## 4. Difficulties in learning hidden structure

## 4.1 Cross-level dependencies

The relative difficulty of learning opaque interactions that we found in our results seems to stem from the fact that the effectiveness of the weightings on each level depends on the weights on the other level. Specifically, high weight on phrase level Faithfulness is only effective when word level constraints are weighted appropriately, while the result of word level weighting can only be transmitted to the surface representation when phrase level Faithfulness has a high enough weight.

We will show here how failure to find appropriate weights on both word level constraints and phrase level Faithfulness simultaneously leads to local optima, and we will show that this scenario is more likely to occur in opaque cases than in transparent cases. Furthermore, we will show how the types of evidence for stratal affiliation of the opaque process offered in the Canadian English datasets increase the likelihood that the learner will find the global optimum.

When the learner has not found weights that generate a desirable distribution at the word level, the learner gets closer to its objective by lowering the weights of phrase level Faithfulness instead of making these weights higher. For instance, consider the weighting below, which leads to a local optimum for opaque 'mitre-cider':

(29) *Sample initialization which leads to local optimum for 'mitre-cider'*

| Word level | | | | Phrase level | | | |
|---|---|---|---|---|---|---|---|
| Ident(son) | Ident(lo) | *V́TV | *aɪ,aʊ/_[-vce] | Ident(son) | Ident(lo) | *V́TV | *aɪ,aʊ/_[-vce] |
| 1 | 7 | 3 | 1 | 0 | 6.28 | 6.29 | 0 |

At the word level, this weighting gives a non-raised diphthong in 'mitre' highest probability, because high-weighted *V́TV and low-weighted *aɪ,aʊ/_[-vce] leads to flapping, which blocks diphthong raising. At the same time, it has flapping and lack of raising on the phrase level, as desired for the opaque interaction (cf. section 3.2). This is illustrated for the word level tableau in (30), and the phrase level tableaux in (31).

(30) *Word level tableau for 'mitre' for the weights in* (29)

| /maɪtɚ/ | Ident(son) 1 | Ident(lo) 7 | *V́TV 3 | *aɪ,aʊ/_[-vce] 1 | H | $e^H$ | p |
|---|---|---|---|---|---|---|---|
| maɪtɚ | | | −1 | −1 | −4 | 0.01 | 0.05 |
| maɪɾɚ | −1 | | | | −1 | 0.37 | 0.95 |
| mʌɪtɚ | | −1 | −1 | | −10 | 0.00 | 0.00 |
| mʌɪɾɚ | −1 | −1 | | | −8 | 0.00 | 0.00 |

(31) *Two of the phrase level tableaux for 'mitre' for the weights in* (29)

| a. mʌɪɾɚ | Ident(son) 0 | Ident(lo) 6.28 | *V́TV 6.29 | *aɪ,aʊ/_[-vce] 0 | H | $e^H$ | p |
|---|---|---|---|---|---|---|---|
| [maɪtɚ] | | −1 | −1 | −1 | −12.57 | 0.000 | 0.00 |
| [maɪɾɚ] | −1 | −1 | | | −6.28 | 0.002 | 0.00 |
| [mʌɪtɚ] | | | −1 | | −6.29 | 0.002 | 0.00 |
| [mʌɪɾɚ] | −1 | | | | 0 | 1 | 1.00 |

| b. maɪɾɚ | Ident(son) 0 | Ident(lo) 6.28 | *V́TV 6.29 | *aɪ,aʊ/_[-vce] 0 | H | $e^H$ | p |
|---|---|---|---|---|---|---|---|
| [maɪtɚ] | −1 | | −1 | −1 | −6.29 | 0.002 | 0.00 |
| [maɪɾɚ] | | | | | 0 | 1 | 1.00 |
| [mʌɪtɚ] | −1 | −1 | −1 | | −12.93 | 0.000 | 0.00 |
| [mʌɪɾɚ] | | −1 | | | −6.28 | 0.002 | 0.00 |

In cases like this, KL-divergence decreases (and fit to the data increases) as the weight of phrase level Ident(low) at the phrase level goes to zero.

(32)
a. *Observed distribution for opaque 'mitre' and 'cider'*

| /maɪtɚ/ | | | | /saɪdɚ/ | | | |
|---|---|---|---|---|---|---|---|
| [maɪtɚ] | [maɪɾɚ] | [mʌɪtɚ] | [mʌɪɾɚ] | [saɪdɚ] | [saɪɾɚ] | [sʌɪdɚ] | [sʌɪɾɚ] |
| 0 | 0 | 0 | 1 | 0 | 1 | 0 | 0 |

b. *KL-divergence for grammars with word level weights as in* (29) *and various weights of phrase level Ident(low)*

| Weights (word level) | | | | /maɪtɚ/ | | | | /saɪdɚ/ | | | | D_KL |
|---|---|---|---|---|---|---|---|---|---|---|---|---|
| Ident (son) | Ident (low) | *V́TV | *aɪ/_ [-vce] | [aɪt] | [aɪɾ] | [ʌɪt] | [ʌɪɾ] | [aɪt] | [aɪɾ] | [ʌɪt] | [ʌɪɾ] | |
| 0 | **6.28** | 6.29 | 0 | 0.00 | 1.00 | 0.00 | **0.00** | 0.00 | **1.00** | 0.00 | 0.00 | 5.87 |
| 0 | **1** | 6.29 | 0 | 0.00 | 0.72 | 0.00 | **0.27** | 0.00 | **0.72** | 0.00 | 0.27 | 1.63 |
| 0 | **0** | 6.29 | 0 | 0.00 | 0.50 | 0.00 | **0.50** | 0.00 | **0.50** | 0.00 | 0.50 | 1.39 |

If the learner has assigned zero weight to phrase level Faithfulness, moving towards appropriate weights at the word level does not lower KL-divergence. This is illustrated in (33) below. Word level and phrase level expected probabilities are shown for 'mitre' only, but KL-divergence is computed for both 'mitre' and 'cider'. Phrase level constraint weights are as given in (29), except for Ident(low), whose weight is set to zero.

(33) *KL-divergence for 'mitre-cider' when phrase level Ident(low) has zero weight*

| Weights | | | | | Word level outputs for /maɪtɚ/ | | | | Surface representations | | | | D_KL |
|---|---|---|---|---|---|---|---|---|---|---|---|---|---|
| Ident (son) Word | Ident (low) Word | *V́TV Word | *aɪ/_ [-vce] Word | Ident (low) Phrase | aɪt | aɪɾ | ʌɪt | ʌɪɾ | [aɪt] | [aɪɾ] | [ʌɪt] | [ʌɪɾ] | |
| 1 | 7 | 3 | 1 | **0** | .05 | .95 | **.00** | .00 | .00 | .50 | .00 | **.50** | 1.39 |
| 4 | 3 | 2 | 4 | **0** | .09 | .64 | **.23** | .03 | .00 | .50 | .00 | **.50** | 1.39 |
| 6 | 3 | 1 | 6 | **0** | .04 | .11 | **.84** | .01 | .00 | .50 | .00 | **.50** | 1.39 |
| 7 | 3 | 0 | 7 | **0** | .02 | .02 | **.96** | .00 | .00 | .50 | .00 | **.50** | 1.39 |

As can be seen in (33), raising the weights of word level Ident(low) and *V́TV and lowering those of word level Ident(sonorant) and *aɪ,aʊ/_[-voice] leads to a desirable result at the word level: raising and no flapping (cf. (23) in section 3.2). However, this information is not factored into the distribution over UR/SR mappings if phrase level Ident(low) has a weight of 0.

At the same time, phrase level Ident(low) has a motivation to decrease its weight until word level outputs with a raised diphthong gain a cumulative probability of at least 0.5. The lower the weight of Ident(low), the closer the winning candidate will be to having 0.5 probability, since setting Ident(low) to zero means that the phrase level will consider raised diphthongs and non-raised ones with equal probability. As long as the current word level weights are such that the winning candidate has less than 0.5 probability, this is an improvement. In the particular case discussed here, Ident(low) will only be motivated to have non-zero weight when the word level gives a boost to raised diphthong outputs, so that the probability of [mʌɪɾɚ] increases as the weight of Ident(low) increases.

Taken together, this means that if the current word level weights lead to lower probability on the attested candidate compared to all-zero weights at the word level, then phrase level Faithfulness might be set to zero before a more desirable set of word level weights might be found.[7] We will call this the bottleneck effect: word level information needs to travel through phrase level Faithfulness constraints in order to have an effect on the distribution over UR/SR mappings. The bottleneck effect is not limited to just this data set, but is applies to any data set with dependence between levels, including other Canadian English data sets (see 4.2 below), and the Southern French case (see 2.3 above).

4.2 Advantage from evidence for stratal affiliation

As we showed in section 3.3, the addition of various kinds of evidence for the stratal affiliation of the opaque process (raising) dramatically increases the likelihood that the opaque interaction will be learned.

Adding 'life' to the opaque interaction means that raising at either the word level or the phrase level will be rewarded, regardless of the weighting of the flapping constraint *V́TV. This means that increasing the weight of word level *aɪ,aʊ/_[-voice] leads to a sharper drop in KL-divergence for opaque 'mitre-cider-life' than for opaque 'mitre-cider-lie-for'[8].

In this manner, the presence of 'life' in the data set means a sharp increase in the gradient of the word level constraint *aɪ,aʊ/_[-voice] in a position in the weight space that would otherwise have a low gradient for that constraint. We have not studied how the exact numerical increase in the gradient of this constraint, which depends on the relative frequency of data points like 'life' in the total corpus (we only considered a minimal corpus here), relates to learnability rates – this is a matter for future research. However, we can confidently say that the categorical presence of raising outside its interaction with flapping has a positive influence on the learnability of the interaction.

(34) *Adding 'life' leads to a stronger effect of representing raising at the word level*

| Word level | | | | Phrase level | | | | $D_{KL}$ | $D_{KL}$ |
|---|---|---|---|---|---|---|---|---|---|
| Ident (son) | Ident (low) | *V́TV | *aɪ,aʊ/ _[-vce] | Ident (son) | Ident (low) | *V́TV | *aɪ,aʊ/ _[-vce] | 'mitre-cider' | 'mitre-cider-life' |
| 1 | 7 | 3 | 1 | 0 | 6.28 | 6.29 | 0 | 5.87 | 11.31 |
| 1 | 7 | 3 | 7 | 0 | 6.28 | 6.29 | 0 | 5.85 | 6.54 |
| 1 | 7 | 3 | 14 | 0 | 6.29 | 6.29 | 0 | 5.85 | 5.85 |

Adding 'lie for' to the opaque interaction means that high weight on phrase level Ident(low) is penalised less strongly when the word level weights do not generate the desirable candidates with high probability[9]. This is because word 'lie' itself contains neither a flapping nor a raising context, so that flapping and raising constraints do not interact with the identity of the diphthong at the word level. It is not necessary to determine the mutual weighting of word level *V́TV and Ident(son) or *aɪ,aʊ/_[-voice] and Ident(low) to give the desirable word level output for 'lie for' more than 0.5 probability – non-zero weight on word level Ident(son) is sufficient.

---

[7] More generally, the effect happens when the current word level weights produce a distribution over surface candidates that has a higher KL-divergence from the actual distribution than a uniform distribution (i.e., when the word level grammar makes it so that predictions for surface forms go in the opposite direction of the actual data).
[8] However, lowering the weight of phrase level Ident(low) is also rewarded more strongly compared to opaque 'mitre-cider', so that the bottleneck effect becomes stronger with the addition of 'life'. This is probably why 'life' only had a modest effect on learnability.
[9] Another effect of adding 'lie for' is that high weight on phrase level *aɪ,aʊ/_[-voice] is penalised more strongly, because this constraint prefers a raised diphthong in 'lie for'.

(35) *Adding 'lie for' makes it less attractive to lower weight on Ident(low)*

| Word level | | | | Phrase level | | | | D$_{KL}$ | D$_{KL}$ |
| --- | --- | --- | --- | --- | --- | --- | --- | --- | --- |
| Ident (son) | Ident (low) | *V́TV | *aɪ,aʊ/ _[-vce] | Ident (son) | Ident (low) | *V́TV | *aɪ,aʊ/ _[-vce] | 'mitre-cider' | 'mitre-cider-lie-for' |
| 1 | 7 | 3 | 1 | 0 | 6.28 | 6.29 | 0 | 5.87 | 5.87 |
| 1 | 7 | 3 | 1 | 0 | 1 | 6.29 | 0 | 1.63 | 1.94 |
| 1 | 7 | 3 | 1 | 0 | 0 | 6.29 | 0 | 1.39 | 2.08 |

When both 'life' and 'lie for' are present in the data set, the grammar must represent the raising process at the word level in order to generate lack of raising in 'lie for' (because word level *aɪ,aʊ/_[-voice] cannot see that 'lie for' has /aɪ/ before /f/, while phrase level *aɪ,aʊ/_[-voice] can). As shown in section 3.2, this introduces an additional dependency between word and phrase level for the transparent datasets: while the transparent interaction between flapping and raising does not require raising to apply before flapping, the combination of 'life' and 'lie for' does.

This creates a bottleneck effect for the transparent dataset, which explains the decrease in learnability for transparent 'mitre-cider-life-lie-for'. The increase in learnability for opaque 'mitre-cider-life-lie-for', on the other hand, can be seen as the cumulative effect of 'life' and 'lie for' on the opaque interaction, as reviewed above.

## 5. Concluding remarks

We have presented here an approach to learning opaque and transparent interactions in a MaxEnt version of Stratal OT. Our goal was to investigate whether the general setup of Stratal OT – chained parallel OT grammars with independent rankings or weightings of constraints – predicts learnability differences between opaque and transparent process interactions, and also whether evidence of a process' stratal affiliation makes it easier to learn opaque interactions.

Our first case study was opaque tensing/laxing in Southern French. We found that it was learned at a high rate of accuracy, but the solution space has a local optimum – one where the grammar does not represent the phonological process at all, which yields free variation in the data.

We then looked at the opaque interaction between diphthong raising and flapping in Canadian English (Joos 1942, Idsardi 2000 and references therein). The opaque raising process also applies in contexts where flapping is irrelevant, and it does not apply across word boundaries – both of which constitute evidence for the stratal affiliation of raising.

We found that, without additional evidence for stratal affiliation, the opaque interaction was learned at a rate of about 50%, while its transparent counterpart was learned at ceiling. However, addition of this additional evidence improved the learnability of the opaque interaction to a maximum of 92%, while the learnability of the transparent interaction descended to a similar rate. This confirms Kiparsky's prediction: evidence of the stratal affiliation of raising does improve its learnability when it is opaque. We also found that the advantage of transparent over opaque interactions is relative: the dataset with full evidence for stratal affiliation did not produce a learnability difference between the two.

Our explanation for the observed learning difficulties in this framework has to do with a bottleneck effect in the transmission of information from earlier derivational levels through Faithfulness constraints. This effect makes it more difficult to find the global optimum when the learner starts with word level weights that predict the wrong surface form. However, this effect is mitigated for opaque Canadian English by information about an opaque process' stratal affiliation, because this information either boosts desirable word level weightings, punishes undesirable phrase level weightings, or diminishes the bottleneck effect in general. See section 4.2 for details.

The two cases that we considered differ in their learnability rate. The Southern French data set contains evidence that the opaque process (tensing/laxing) is word-bounded, making it analogous to the opaque 'mitre-cider-lie-for' data set in the Canadian English case study. While Southern French was learned at a rate of 96%, 'mitre-cider-lie-for' was only learned at a rate of 87%. A possible explanation is that the Southern French case study did not include constraints on syllable structure, while the process that makes tensing/laxing opaque is resyllabification. We predict that the learnability of Southern French would go down slightly if such constraints were introduced into the simulation.

We limited ourselves here to two case studies and one learning implementation. Other approaches to learning Stratal OT are possible too: probabilistic ranked constraint learning with Expectation Maximization (Jarosz 2006, 2016, ms), Noisy Harmonic Grammar (Coetzee & Pater 2011), Stochastic OT (Boersma 1998). These might differ in their particular learning strategies, the local optima they find, and the distribution over outputs generated at local optima. Nonetheless, the mechanisms responsible for the learnability differences that we found are quite general. Interdependence between phrase level Faithfulness and appropriate weighting at the word level poses a challenge for finding a grammar that generates the learning data. For this reason, we predict that other learning approaches will find results similar to ours (an early confirmation of this can be found in Jarosz 2016). However, more work is needed that explores the learnability of these and other opaque interactions in other learning frameworks. Further work is also needed to examine how other cases of opacity behave in our framework. In particular, more complex interactions that involve more constraints and/or more derivational levels would be essential to test the predictions of a general advantage for transparency, and of improved learning for opacity given evidence of stratal affiliation. In addition, as pointed out in section 4.2, it is important to understand how prevalent evidence for stratal affiliation must be in the data to ensure successful learning.

Finally, further research would need to be done before one could draw conclusions about the relationship of our findings to attested patterns of language learning, language change, and language typology. In naturalistic language learning, we are unaware of any evidence that learners have biases toward transparent interactions, or that they find some cases of opacity easier than others. Such evidence may be difficult to come by outside of the laboratory, since it is general to control for confounds when comparing learning of different patterns. There is some evidence that transparent interactions are easier in artificial language learning (Kim in press), but this line of research is only beginning. There is also evidence that opacity can be innovated by children (e.g. Dinnsen & Farris-Trimble 2008); whether or not this is a challenge to our model, or other related theories, would require a detailed consideration of the constraints involved and the steps on the learning path. To apply our model to historical change would require adding the analogue of rule loss, that is the possibility of mis-learning an opaque interaction as a contrast; our constraint set cannot generate this possibility. And to make conclusions about how these results fit the relative degree of typological attestation of opaque and transparent interactions, we would need not only a model that connects learning to typological attestation (see e.g. Staubs 2014a), but also empirical research to determine whether opacity is underattested or not.